\documentclass{article}

\usepackage[english]{babel}

\usepackage[letterpaper,top=2cm,bottom=2cm,left=3cm,right=3cm,marginparwidth=1.75cm]{geometry}
\usepackage{amsmath}
\usepackage{graphicx}

\usepackage[colorlinks=true, allcolors=blue]{hyperref}
\title{Agent-E: From Autonomous Web Navigation to Foundational Design Principles in Agentic Systems}
\author{%
Tamer Abuelsaad, 
Deepak Akkil, 
Prasenjit Dey, 
Ashish Jagmohan, \\
Aditya Vempaty, 
Ravi Kokku
}

\date{\{tea, deepak, prasenjit, ashish, aditya, ravi\}@emergence.ai\\
\vspace{0.1in}
Emergence AI \\
\vspace{0.1in}
July, 2024}

\begin{document}
\sloppy
\maketitle
\begin{abstract}
AI Agents are changing the way work gets done, both in consumer and enterprise domains. However, the design patterns and architectures to build highly capable agents or multi-agent systems are still developing, and the understanding of the implication of various design choices and algorithms is still evolving. 
%and their implications are not so well understood.
In this paper, we present our work on building a novel web agent, Agent-E \footnote{Our code is available at \url{https://github.com/EmergenceAI/Agent-E}}. Agent-E introduces numerous architectural improvements over prior state-of-the-art web agents such
as hierarchical architecture, flexible DOM distillation and denoising method, and the concept of \textit{change observation} to guide the agent towards more accurate performance.
We first present the results of an evaluation of Agent-E on WebVoyager benchmark dataset and show that Agent-E beats other SOTA text and multi-modal web agents on this benchmark in most categories by 10-30\%. We then synthesize our learnings from the development of Agent-E into general design principles for developing agentic systems. These include the use of domain-specific primitive skills, the importance of distillation and de-noising of environmental observations, the advantages of a hierarchical architecture, and the role of agentic self-improvement to enhance agent efficiency and efficacy as the agent gathers experience. 
\end{abstract}

\section{Introduction} \label{sec:intro}

Recent advancements in large language models have led to significant interest in the development of autonomous agents that can execute complex tasks on the web \cite{zheng2024gpt, he2024webvoyager, lutz2024wilbur} and on device \cite{bai2024digirl, wen2024autodroid}. Automation of complex and repetitive tasks presents an invaluable opportunity to increase individual and organizational efficiency. In addition, robust automation systems will unlock new use cases and experiences by enabling collaboration between users and AI agents working together to accomplish sophisticated tasks.

\begin{figure} [h!]
    \centering
    \includegraphics[width=0.75\linewidth]{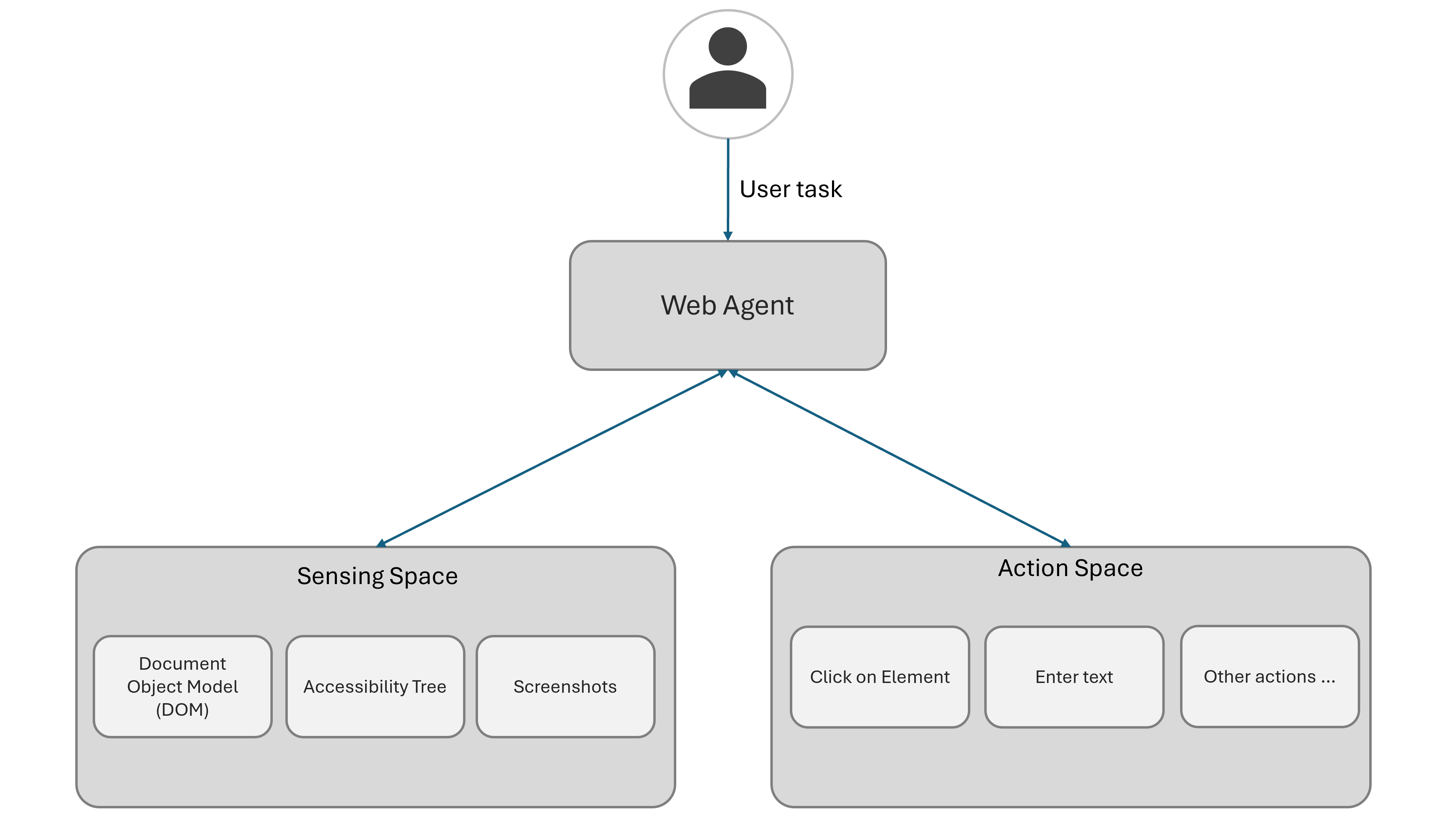}
    \caption{Simplified anatomy of web agents}
    \label{fig:agentAnatomy}
\end{figure}

Figure \ref{fig:agentAnatomy} shows a simplified anatomy of a web agent. A typical web-agent, at its most basic, has two key capabilities: the capability to sense the web page and the capability to act on the web page. 
\begin{itemize}
 \item Sensing: Sensing the state of the web-page, for a web agent, typically involves encoding the Document Object Model (DOM) of the page \cite{nakano2022webgpt, lutz2024wilbur}, using the accessibility tree of the page \cite{he2024webvoyager} and/or using screenshots of the page \cite{he2024webvoyager}. 
 \item Acting: The action space can be comprised of simple actions such as navigating to URLs, clicking on elements, and entering text in a field, or composite actions comprising of several simple actions. An example of composite action is used by Lutz et al. \cite{lutz2024wilbur}, which they call `Input'. `Input' selects a text box, deletes any existing content, inputs text and presses submit button.
\end{itemize}
Based on the user's task and the current state of the page, the agent can create a plan i.e., it determines the subsequent next action or series of actions required to complete the task.

Developing a robust web agent that can autonomously perform tasks on the browser presents multiple challenges. Firstly,  HyperText Markup Language (HTML) Document Object Models (DOMs)  can be noisy and expansive, so that they often exceed an LLM’s context window. This renders them unusable without intelligent simplification or de-noising steps. Further, even with simplification, information may exceed the context window limit of modern LLMs after a few interactions. This is an important challenge that restricts their use for complex tasks \cite{he2024webvoyager}. Secondly, websites are primarily designed for human visual consumption, wherein information is organized to prevent information overload for a user; thus, websites utilize established human-computer interaction patterns that may not be optimal for agent interaction. Complex widgets, such as date selectors, are easy for humans to operate but pose difficulties for agents \cite{lutz2024wilbur}. Thirdly, while humans can naturally execute complex web tasks (e.g., finding the cheapest flight between two destinations), agents require detailed multi-step planning to execute such tasks. Further, the plan may need multiple revisions if the initial approach proves unfruitful. 

Despite these challenges, recent work has demonstrated promising outcomes with web agents executing diverse tasks on the internet. Yet, state-of-the-art web agents leave much to be desired in terms of practical usefulness \cite{wornow2024eclair}. Most previous studies simply report success rates. Task success rates show that web agents are more error-prone compared to a human performing the same task and are not yet ready for effective mainstream use \cite{lutz2024wilbur, he2024webvoyager, zhou2023webarena}.  

Task success rates, while important, may not tell the whole story when it comes to evaluating web-agents. For example Kapoor et al. \cite{kapoor2024ai} note that since that the underlying LLMs are stochastic systems, simple retry mechanisms often improve success rates. Thus, simply measuring task success rates is misleading. There are other important factors that influence the utility and user experience of using agentic systems, such as:
\begin{itemize}
\item Task completion times: The duration required for the system to complete a task, which impacts overall efficiency and productivity. Task completion times are especially relevant for online tasks or tasks with a human in the loop. The task completion times can be influenced by extrinsic factors (network speed, computing environment influencing loading times of web pages) or intrinsic factors (token generation speed of the LLMs, the path taken by the agent to perform the task, errors made during execution etc). Despite the variability, task completion times is an important measure since it can be critical in many real world applications.

\item Cost: The total expenditure involved in executing the task. The cost is a variable measure and depends on the underlying model and the per token input/output cost of different model providers. This is also a `point in time' measure, since the pricing models set by LLM providers change frequently. While not fully encompassing, a good proxy for cost is number of LLM calls involved. 

\item Error-awareness: The capability of the system to recognize and report its own errors or failures, ensuring reliability and ability to provide a seamless user experience by highlighting and recommending fallback options. Ability of system to detect that it has not completed the task appropriately will also open up avenues to collect human demonstration (e.g. \textit{I could not complete this task, can you demonstrate how you would do it?}) and a path towards continuous improvement. 
\end{itemize}

In this paper, we introduce Agent-E, a state-of-the-art web agent capable of performing complex web-based tasks. Central to Agent-E are two LLM-powered components: the planner agent and the browser navigation agent. The planner agent is responsible for task planning and task management. It breaks down the user task into a sequence of sub tasks and delegates them one at a time to the browser navigation agent. The browser navigation agent is tasked with executing the individual sub tasks by sensing the page using different DOM distillation capabilities available to it, finding the next actions to execute and reporting its task success or failure back to the planner. This loop is executed iteratively until the task is completed. This tiered architecture ensures that the planner agent is insulated from the overwhelming and noisy details of the website and DOM, and the browser navigation agent is freed from the complexities of the overall task planning and orchestration. 

Agent-E can run in two modes: autonomous mode and human-in-the-loop mode. In the autonomous mode, Agent-E performs the task end to end as best as it can, and in the human-in-the-loop mode, it falls back to the user when it encounters steps that it cannot accomplish (e.g. logging to a web page, solve a captcha triggered by the website) or to ask for clarifications when the task itself is ambiguous. We believe human-in-the-loop workflows are critical for adoption of agentic systems. However, since there are no web automation benchmarks for human-in-the-loop agents \cite{kapoor2024ai}, in this paper we will focus on Agent-E as an autonomous web agent.

We evaluate Agent-E in autonomous mode on the WebVoyager benchmark \cite{he2024webvoyager}, where we achieve a new state-of-the-art result of 73.2\%, 20\% higher than previous state-of-the-art for text-only \cite{lutz2024wilbur} and 16\% higher than previous state-of-the-art multi-modal web agent \cite{he2024webvoyager}. For specific sites like Wolfram Alpha, it achieves upto 30\% improvement.

\subsection{Contributions}
\begin{itemize}
\item We introduce a novel hierarchical architecture for web agents that enables the execution of more complex tasks through a clear separation of roles and responsibilities between a planner agent and a browser navigation agent. We illustrate the utility of the hierarchical architecture for capabilities such as task verification, backtracking, and error recovery, with examples where these capabilities are invoked.

\item We propose a flexible DOM distillation approach whereby, given a task, the browser navigation agent can choose the most suitable DOM representation from three different implementations. We demonstrate its effectiveness with examples.

\item We highlight the benefits of \textit{`Change observation'}, a paradigm conceptually similar to Reflexion \cite{shinn2024reflexion}. After the execution of each action, the outcome (change in state) is monitored and used to provide verbal feedback to the browser navigation agent. This guides the agent towards better awareness of the current state and more accurate performance.

\item We report detailed end-to-end evaluations of Agent-E on the WebVoyager benchmark and show that it achieves a new state-of-the-art results with a 73.2\% success rate. This marks a 20\% improvement over the previous text-only web agent and a 16\% increase over the former state-of-the-art multi-modal web agent. 

\item Beyond task success rates, we are the first to report additional metrics such as error awareness, task completion times, and the number of LLM calls, offering a baseline for comprehensive evaluation of future web agents.

\item Agent-E is a result of numerous experimentation and exploration into the space of AI Agents. We synthesize our learning from the development of Agent-E into design principles for development of agentic systems that we believe generalize beyond Agent-E and web automation, and provide useful insights to practitioners.

%\item (To be removed?: )Based on the comprehensive evaluation, we present a new architecture for Agent-E 2.0,  which we believe to be significantly closer to mainstream use. AJ: agreed - we can update the arxiv page with this when we have results

\end{itemize}

\section{Agent-E:  System Description}
\begin{figure} [h!]
    \centering
    \includegraphics[width=0.85\linewidth]{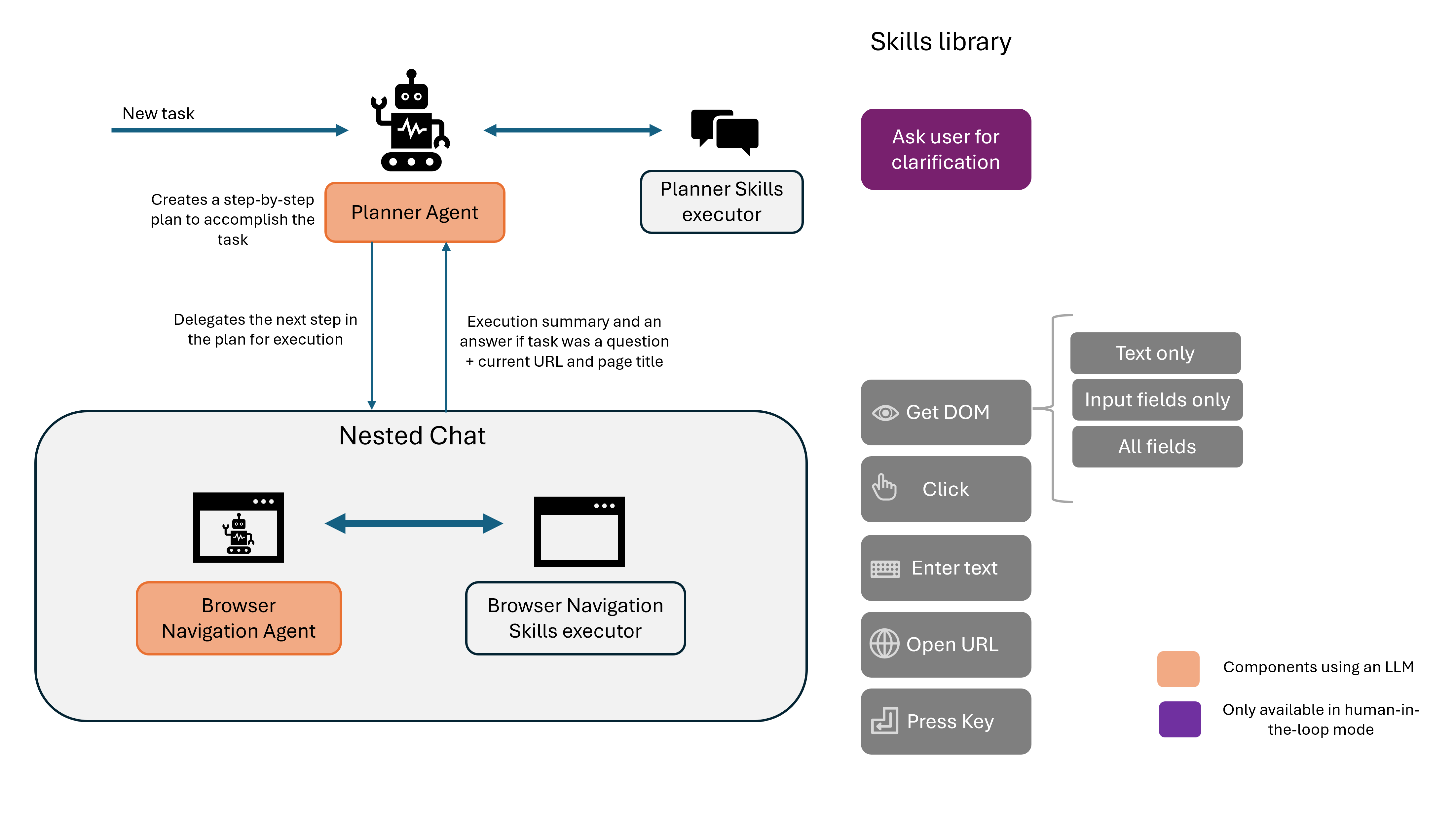}
    \caption{A high level architecture of Agent-E}
    \label{fig:agente-architecture}
\end{figure}

Figure \ref{fig:agente-architecture} shows the high level architecture of Agent-E. It comprises of two LLM-powered agents: Planner Agent and Browser Navigation Agent, and two executor components: Planner Skills executor and Browser Navigation Skills executor. Each LLM-powered agent has skills associated with it, which are python functions that are described to the LLM for function calling. We do not make any distinction between skills associated with sensing (e.g. \emph{Get DOM}) and skills associated with acting on the page (e.g. \emph{click on element}). The executor components execute the function suggested by the LLM and relays the response back to the LLM.  

Agent-E is built using Autogen, the open-source programming framework for building multi-agent collaborative systems \cite{wu2023autogen} and Playwright\footnote{\url{https://playwright.dev/}} for browser control. Its architecture leverages the interplay between skills and agents as shown in Figure \ref{fig:agente-architecture}. 

\begin{figure} [h!]
    \centering
    \includegraphics[clip, trim=0 0 0 0, width=\linewidth]{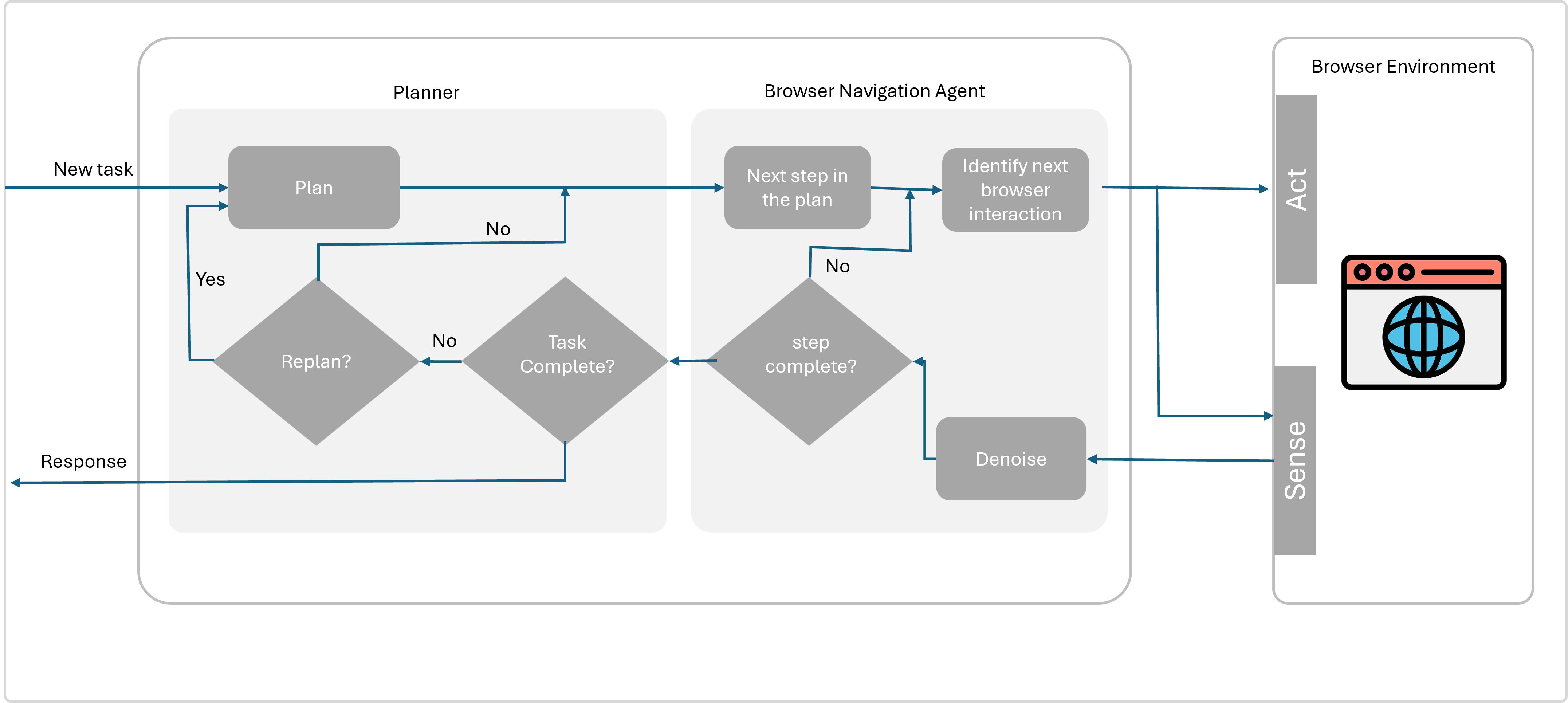}
    \caption{Conceptual flow diagram of Agent-E. Individual blocks represent functions; In Agent-E, a single LLM call is used to perform multiple functions.}
    \label{fig:agente-flow}
\end{figure}

\begin{figure} [h!]
    \centering
    \includegraphics[width=1\linewidth]{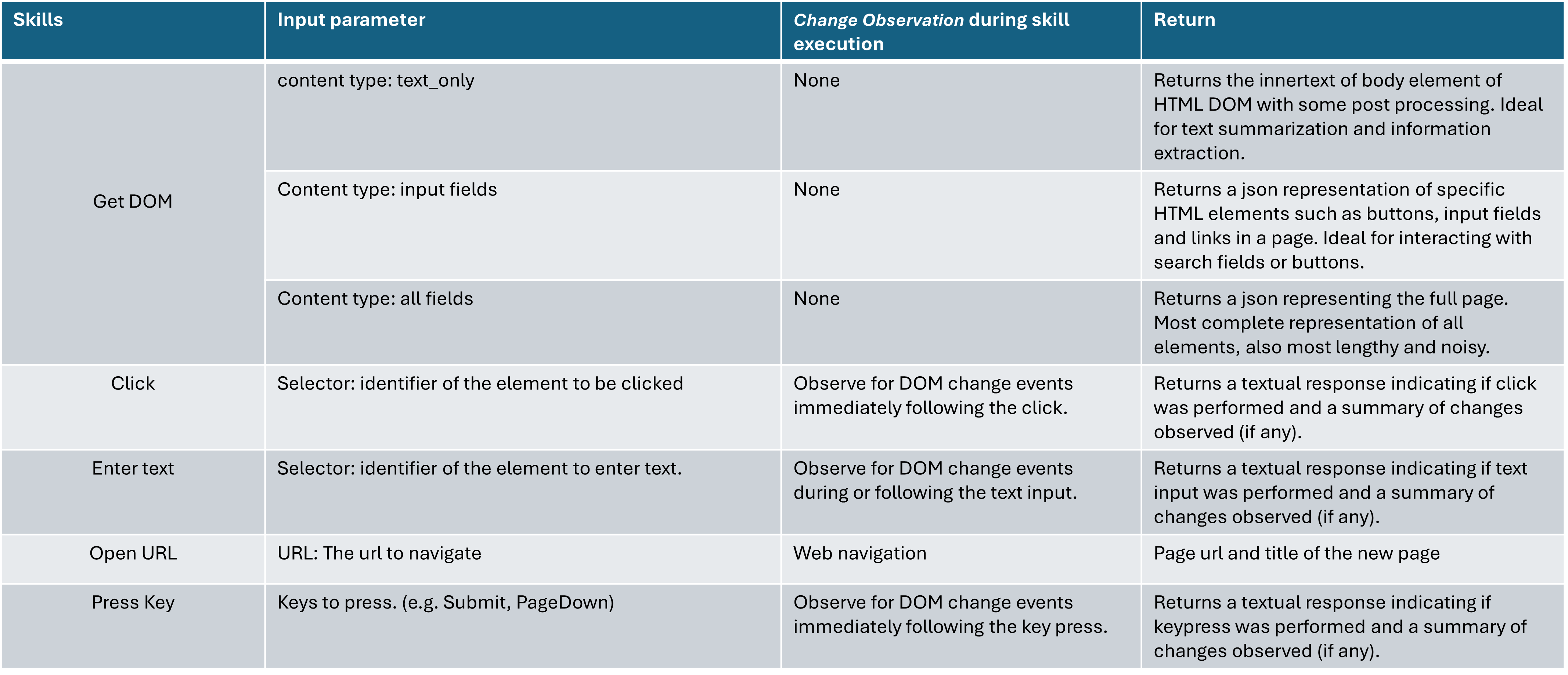}
    \caption{Skills registered to the Browser Navigation Agent for sensing and acting on the web page.}
    \label{fig:agente-skills}
\end{figure}

\begin{figure} [h!]
    \centering
    \includegraphics[clip, trim=0 0 0 0, width=\linewidth]{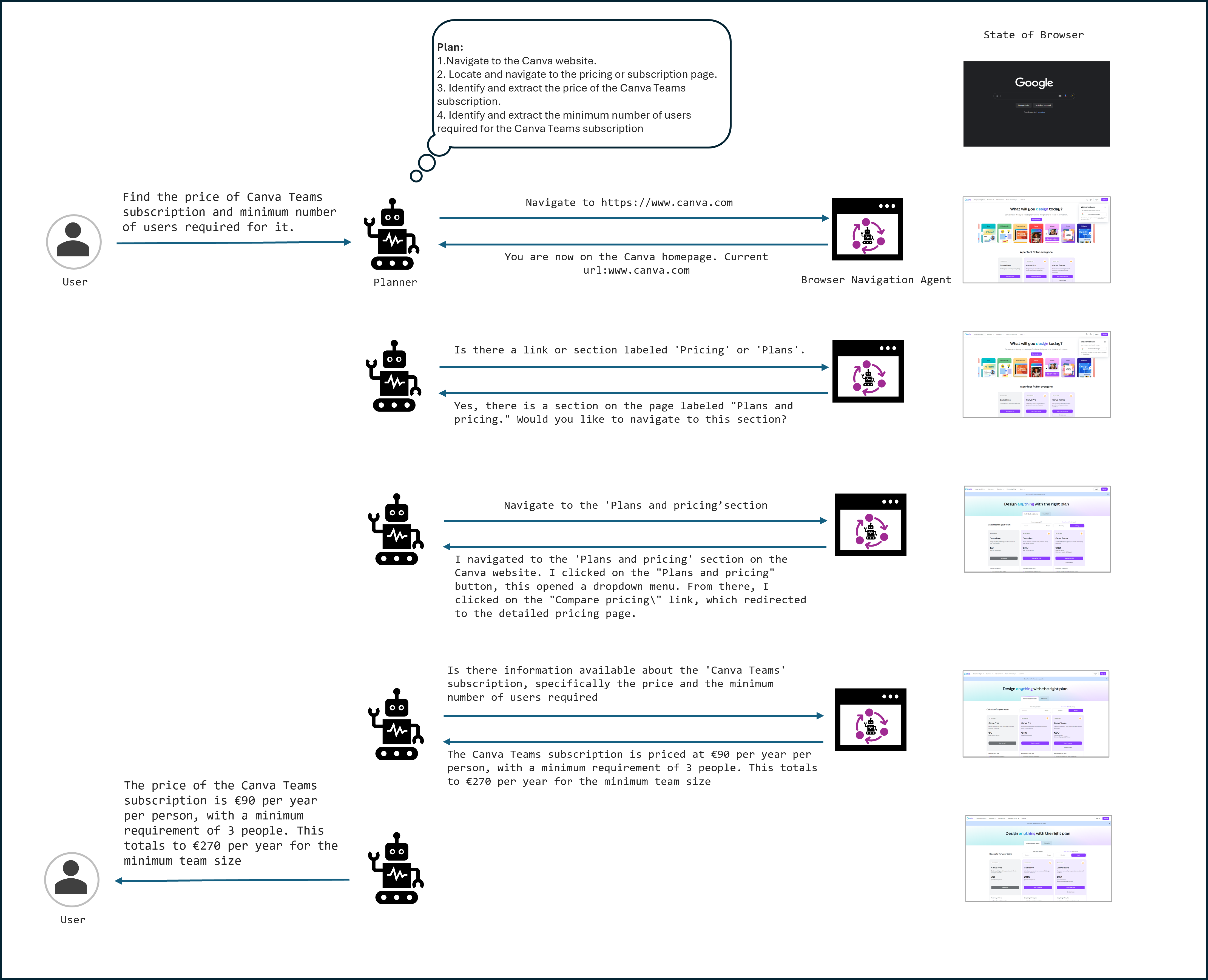}
    \caption{An example of Agent-E execution highlighting communication between the planner and browser navigation agent for the user task \textit{Find the price of Canva Teams subscription and minimum number of users required for it}}
    \label{fig:agente-planner}
\end{figure}

Figure \ref{fig:agente-flow} shows a conceptual flow diagram of Agent-E. Given a new user task, the planner decomposes the task into a sequence of steps that need to be executed. The next step is then delegated to the browser navigation agent for execution. Browser navigation agent is implemented as a nested chat that is freshly instantiated by the Planner for each run (i.e., it does not contain previous steps in its chat history). Browser navigation agent has a set of predefined `primitive' or foundational skills for observing denoised browser state and controlling the browser instance using Playwright. Figure  \ref{fig:agente-skills} shows the skills available to the browser navigation agent.  The browser navigation agent uses the skills available to it to perform the sub task and return a summary of actions it took to perform the task and/or answer the planner if the task was a question. Figure \ref{fig:agente-planner} shows an example communication between planner and the browser navigation agent and Figure \ref{fig:agente-nested} shows the browser navigation agent using the primitive skills available to it to perform a related sub task.

\subsection{Skills Design for Browser Navigation Agent}

\begin{figure} [h!]
    \centering
    \includegraphics[clip, trim=0 0 0 0, width=\linewidth]{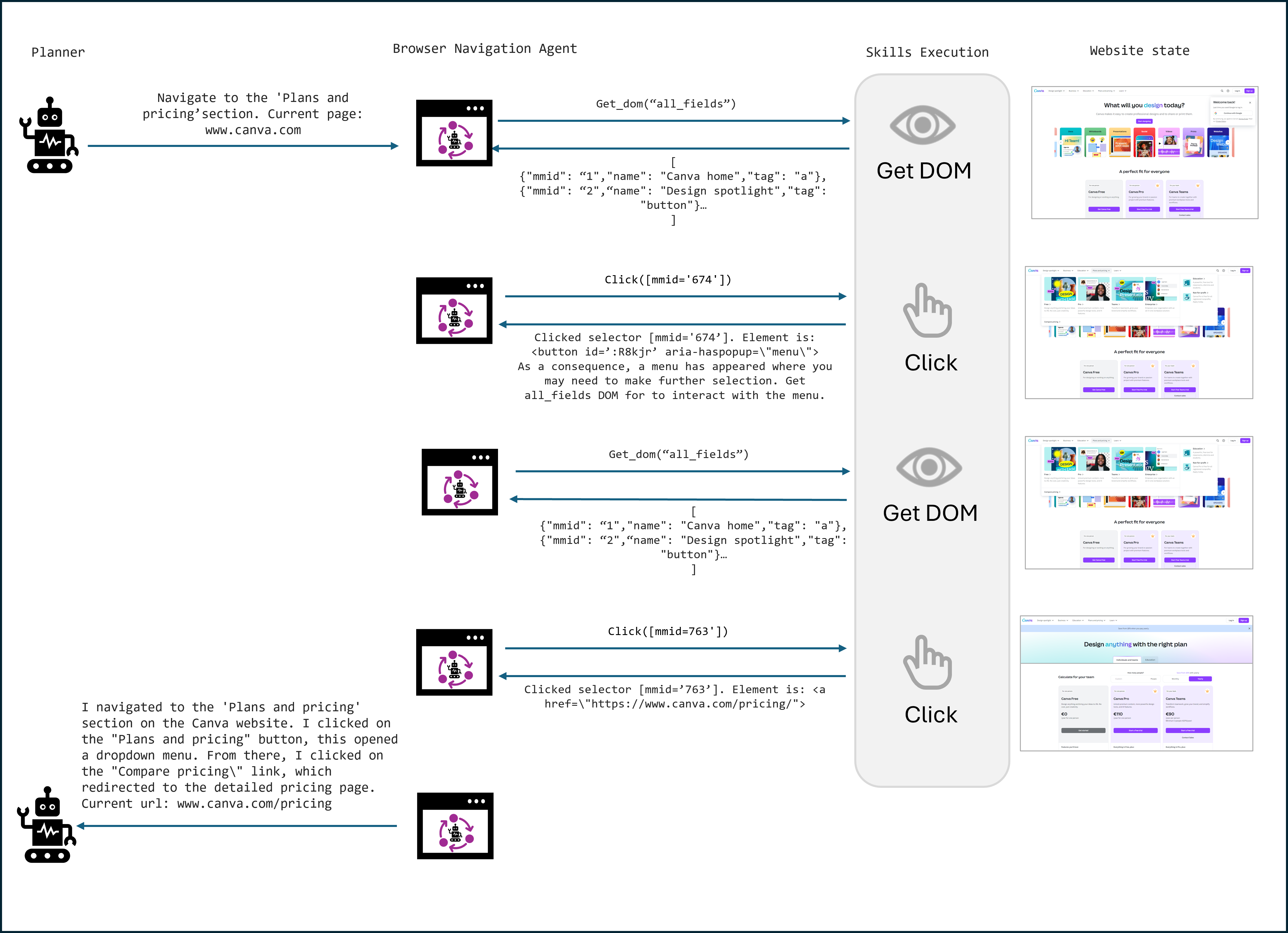}
    \caption{An example of Agent-E nested chat execution loop for the sub task \textit{``Navigate to the plans and pricing section"} which is part of the larger task introduced earlier \textit{``Find the price of Canva Teams subscription and minimum number of users required for it"}}
    \label{fig:agente-nested}
\end{figure}

There are two key differences between Agent-E and previous web agent implementations such as Wilbur \cite{lutz2024wilbur} and WebVoyager \cite{he2024webvoyager} in terms of skills .
\begin{itemize}
\item  Sensing Skills: Agent-E supports multiple DOM synthesis techniques that allows the browser navigation agent to choose the approach best suited for the task (see Figure \ref{fig:agente-skills}). If the task is to summarise information on a page, it can simply use Get DOM with \textit{text\_only} content type. If the task is to identify and execute a search on a page, it can use the content type \textit{input\_fields}. If the task is to list all the interactive elements on a page, it can use \textit{all\_fields}. This optimizes the information available to the agent and prevents the problems associated with noisy DOM. Another key difference is our DOM de-noising techniques for \emph{all\_fields} and \emph{input\_fields} attempts to preserve the parent child relationship of elements wherever possible and relevant. This is unlike some of the other previous implementations which uses a flat DOM encoding (e.g. \cite{lutz2024wilbur}). Further, to make identifying and interacting with HTML elements easier, Agent-E injects a custom identifier attribute (\textit{mmid}) to each element as part of sensing, similar to \cite{zhou2023webarena} and \cite{he2024webvoyager}.  

\item Action Skills: All the action skills are designed to not only execute an action but also report on any change in state as an outcome of the action, a concept we call \textit{`Change observation'}. This is conceptually similar to Reflexion paradigm \cite{shinn2024reflexion} which uses verbal reinforcement to help agents learn from prior failings. However, a key difference is that change observation is not directly associated or limited to a prior failure. The observation returned can be any type of outcome of the action. For example, a click action may return a response \textit{Clicked the element with mmid 25. As a consequence, a popup has appeared with following elements}. Such verbal responses nudges the agent towards the correct next step.

\textit{Change Observation} capability is implemented by observing changes in selected attributes of elements (e.g \textit{aria-expanded}) and Mutation Observer Web API \footnote{\url{https://developer.mozilla.org/en-US/docs/Web/API/MutationObserver}} that allow observing changes in DOM immediately following an action execution. This is especially useful for extremely dynamic pages such as a flight booking website and nudges the LLM to take appropriate next step given the task. (see Figure \ref{fig:agente-nested} for an example)
\end{itemize}

\section{Evaluation}
\subsection{Introduction to WebVoyager}
\begin{figure} [h!]
    \centering
    \includegraphics[width=0.5\linewidth]{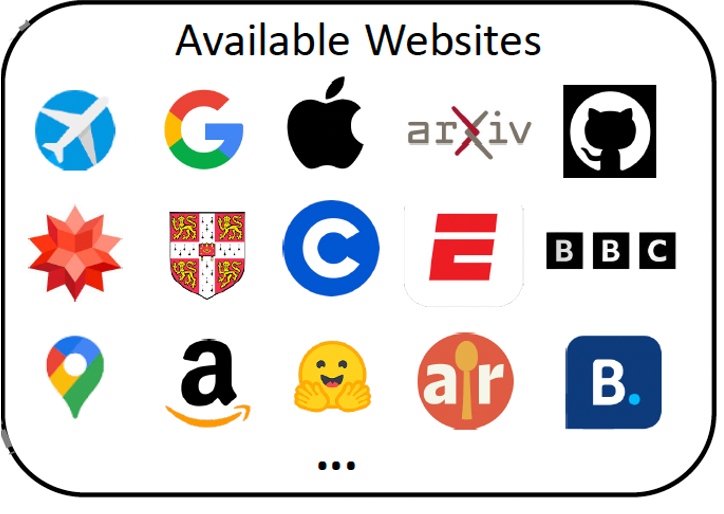}
    \caption{WebVoyager Websites}
    \label{fig:web_voyager_websites}
\end{figure}

WebVoyager \cite{he2024webvoyager} is a recent web agent benchmark that consists of web navigation tasks across 15 real websites as shown Figure  \ref{fig:web_voyager_websites}. Each website has about 40-46 tasks resulting in a benchmark dataset of 643 tasks (see Table \ref{table:exampletasks} for example tasks). These tasks could be completed through DOM manipulation (textual) as well as augmenting with image understanding (multi-modal). We chose WebVoyager since it covers a diverse range of tasks across different live, dynamic and representative websites. Other alternative benchmarks either focused solely on single task domain \cite{yao2022webshop}, used custom created websites that are significantly simpler than real-world versions in terms of DOM complexity \cite{zhou2023webarena}, or used cached versions of real-websites only supporting a fixed route for the agent and not allowing free form exploration \cite{deng2024mind2web}. 
One of the issues with WebVoyager though is that it uses static dates for some of the tasks (e.g. \textit{Search a hotel with free WiFi and air conditioning in Bali from Jan 1 to Jan 4, 2024.}). The static dates makes some of these tasks unachievable since the dates have passed. We manually went through the tasks to identify ones with fixed dates in the tasks that make the task unachievable, and added 8 months to the dates (i.e., \textit{Search a hotel with free WiFi and air conditioning in Bali from Aug 1 to Aug 4, 2024.}).

\begin{table}[h!]
\begin{tabular}{|p{2cm}|p{13cm}|}
\hline
Website &  Task \\ 
\hline
Allrecipes &  Find a recipe for Baked Salmon that takes less than 30 minutes to prepare and has at least 4-star rating based on user reviews. \\ 
\hline
Booking & Find the cheapest available hotel room for a three-night stay from 1st Jan in Jakarta. The room is for 2 adults, just answer the cheapest hotel room and the price.  \\
\hline
Github & Search for an open-source project related to `climate change data visualization' on Github and the report the project with the most stars. \\
\hline
Google Flights & Search a one-way flight from Dublin to Athens Greece for 1 Adult that leaves on December 30 and analyze the price graph for the next 2 months.\\
\hline
\end{tabular}
\caption{Example tasks from WebVoyager benchmark.}
\label{table:exampletasks}
\end{table}

\subsection{Evaluation Setup}
The entire benchmark was divided among 5 human evaluators who ran 125-130 tasks each. For each task, the evaluators were instructed to classify the task as \textit{pass} or \textit{fail} along with a textual reason in case of failures. They were instructed to mark the task as pass if it was completed successfully in full (in case the task has multiple parts). The evaluators ran the task from India during IST office hours (this may have implications when you interpret or compare measures such as task completion times). 
For the evaluation, Agent-E was used in autonomous mode (i.e., \emph{ask user for input} skill was disabled) and used GPT-4-Turbo as the LLM for both planner and browser navigation agent.

\subsection{Measures}
We report four important measures that are relevant for comprehensive evaluation of web agent and understanding their practical implementation readiness.
\begin{itemize}
\item Task success rates: The percentage of tasks that Agent-E performed successfully across websites.
\item Self-aware vs Oblivious failure rates: Detecting when the task was not completed successfully is of utmost importance, since it can be used for enabling a fallback workflow, to notify the user of failure or use as an avenue to gather human demonstration for the same task. 
Self-aware failures are failures where agent is aware of its own failure in completing the task and responds with a final message explicitly stating so, e.g. \textit{I'm unable to provide a description of the first picture due to limitations in accessing or analyzing visual content.} or \textit{`Due to repeated rate limit errors on GitHub while attempting to refine the search...'}. The failures could be due to technical reasons or agent deeming the task unachievable since it could not complete the task after repeated attempts. On the other hand, oblivious failures are cases were the agent wrongly answers the question or performs the wrong action (e.g. adds the wrong  product to cart or provides a wrong information). For mainstream utility, oblivious failures should be as minimal as possible.  
For the current evaluation, failures were categorized to self-aware and oblivious failures by manual annotation. However, it would be trivial to employ an LLM critique to automatically do the same task, similar to \cite{wornow2024eclair}.

\item Task completion times: The average time required to complete the task, across websites for failed and successful tasks.
\item Total number of LLM calls:  The average number of LLM calls (both planner and browser navigation agent) that was required to perform the task. This includes both successful and failure cases.
\end{itemize}

\subsection{Result: Quantitative}

\begin{table}[h!]
\centering
\begin{tabular}{|c c c c c c c c c |} 
 \hline
  Publication & \multicolumn{8}{|c|}{Task success rates on websites} \\ [0.2ex] 
   \hline
   & Allrecipe & Amazon & Apple & Arxiv &  Github  &  Booking  & ESPN & Coursera\\ 
 \hline \hline 

 He et al. 2024 (text) & 57.8 & 43.1 & 36.4 & 50.4 & 63.4 & 2.3 & 28.6 & 24.6\\ 
 He et al. 2024 (multi) & 51.1 & 52.9 & 62.8 & 52.0 & 59.3 & 32.6 & 47.0 & 57.9\\ 
 Lutz et al. 2024 (text) & 60 & 43.9 & 60.5 & 51.2 & 22.0 & \textbf{38.6} & 59.1 & 51.1\\ 
 \textbf{Agent-E (text)} & \textbf{71.1} & \textbf{70.7} & \textbf{74.4} & \textbf{62.8} & \textbf{82.9} & 27.3 & \textbf{77.3} & \textbf{85.7} \\ 
 \hline \hline 

 Failure modes & \multicolumn{8}{|c|}{Agent-E Error Analysis on Websites} \\ [0.2ex] 
 \hline
Overall failures \%  & 28.9 & 29.3 & 25.6 & 37.2 & 17.1 & 72.7 & 22.7 & 14.3 \\ 
Self-aware failures \%  & 17.8 & 14.6 & 9.3 & 18.6 & 12.2 & 4.5 & 13.6 & 4.8 \\ 
Oblivious failures \% & 11.1 & 14.6 & 16.3 & 18.6 & 4.9 & 68.2 & 9.1 & 9.5 \\ 
 \hline \hline
  TCT & \multicolumn{8}{|c|}{Agent-E Avg. Task Completion Times (seconds)} \\ [0.2ex] 
 \hline
TCT (Success) & 116 & 286 & 122 & 137 & 104 & 183 & 187 & 119 \\ 
TCT (Failed) & 196 & 246 & 200 & 176 & 384 & 317 & 387 & 177 \\ 
\hline 
 \hline \hline
  LLM Calls & \multicolumn{8}{|c|}{Agent-E Avg. Number of LLM calls} \\ [0.2ex] 
 \hline
Total & 22 & 23.1 & 21.5 & 25.5 & 21.5 & 36.4 & 24.0 & 25.5 \\ 
Planner & 6.5 & 6.4 & 5.9 & 6.9 & 5.4 & 6.6 & 6.3 & 6.3 \\
Browser Nav Agent & 15.5 & 16.7 & 15.6 & 18.6 & 16.1 & 29.8 & 17.7 & 19.2 \\
\hline
\end{tabular}
\caption{Evaluation of Agent-E on WebVoyager.}
\label{table:results1}
\end{table}

\begin{table}[h!]
\centering
\begin{tabular}{|c c c c c c c c c |} 
 \hline
  Publication & \multicolumn{8}{|c|}{Task success rates on websites} \\ [0.2ex] 
   \hline
   & Dictionary & BBC & Flights & Maps & Search & Hug.Face & Wolfram & Overall\\ 
 \hline \hline 

 He et al. 2024 (text) & 66.7 & 45.2 & 7.1 & 62.6 & 75.2 & 31.0 & 60.2 & 44.3\\ 
 He et al. 2024 (multi) & 71.3 & 60.3 & \textbf{51.6} & 64.3 & 77.5 & 55.8 & 60.9 & 57.1\\ 
 Lutz et al. 2024 (text) & \textbf{86.0} & \textbf{81.0} & 0.0 & 39.0 & 67.4 & 53.5 & 65.2 & 52.6\\ 
 \textbf{Agent-E (text)} & 81.4 & 73.8 & 35.7 & \textbf{87.8} & \textbf{90.7} & \textbf{81.0} & \textbf{95.7} & \textbf{73.1} \\ 
 \hline \hline 

 Failure modes & \multicolumn{8}{|c|}{Agent-E Error Analysis on Websites} \\ [0.2ex] 
 \hline
Overall failures \%  & 18.6 & 26.2 & 64.3 & 12.2 & 9.3 & 19.0 & 4.3 & 26.9 \\ 
Self-aware failures \%  & 16.2 & 9.6 & 57.1 & 12.0 & 4.6 & 14.3 & 2.1 & 14.1 \\ 
Oblivious failures \% & 2.4 & 16.6 & 7.1 & 0 & 4.6 & 4.7 & 2.1 & 12.8 \\ 
 \hline \hline
 TCT & \multicolumn{8}{|c|}{Agent-E Avg. Task Completion Times (seconds)} \\ [0.2ex] 
 \hline
TCT (Success) & 98 & 105 & 244 & 127 & 106 & 140 & 68 & 150 \\ 
TCT (Failed) & 136 & 110 & 234 & 177 & 135 & 167 & 94 & 220 \\ 
\hline 
\hline \hline
LLM Calls & \multicolumn{8}{|c|}{Agent-E Avg. Number of LLM calls per Task} \\ [0.2ex] 
 \hline
Total & 22.0 & 21.3 & 53.8 & 22.9 & 19.4 & 22.8 & 14.5 & 25.0 \\ 
Planner & 6.6 & 6.0 & 11.4 & 5.8 & 5.6 & 6.2 & 4.4 & 6.4 \\
Browser Nav Agent & 15.4 & 15.3 & 42.2 & 17.0 & 13.7 & 16.6 & 10.15 & 18.6 \\
\hline
\end{tabular}
\caption{Evaluation of Agent-E on WebVoyager (Contd.)}
\label{table:results2}
\end{table}

In this section, we will present quantitative results of how well Agent-E performs in WebVoyager benchmark. Tables \ref{table:results1} and \ref{table:results2} shows the summary of evaluation of Agent-E on WebVoyager. 
\subsubsection*{Comparison with prior state-of-the-art agents}
Agent-E overall successfully completed 73\% of the tasks and outperforms prior state-of-the-art text-only web agent WILBUR \cite{lutz2024wilbur} by 21\%  and state-of-the-art multi-modal web agent \cite{he2024webvoyager} by 16\%. The effectiveness of Agent-E varied across websites from 27.3\% (Booking) to 95.7\% (WolframAlpha). Comparing by websites, Agent-E outperformed both Wilbur \cite{lutz2024wilbur} and WebVoyager multi-modal agent \cite{he2024webvoyager} in 11 out of 15 websites. Booking.com and Google Flights continue to be websites, where all web agents, including Agent-E does not do well. 

It is important to note that \cite{lutz2024wilbur} uses task and website-specific prompting and \cite{he2024webvoyager} uses vision for observing the page. In contrast, Agent-E is a generic text-only web agent which does not employ any task or website-specific instructions. This suggests that there is likely room for further improvement in Agent-E using some of these strategies.

\subsubsection*{Agent-E failure modes}
Overall, Agent-E was self-aware of the failures for more than 52\% of the failed tasks. Typically, Self-aware failures occur when the reason for failure are technical in nature (e.g., navigation issues, inability to extract certain information from DOM elements such as Iframes, canvas or images, inability to operate a button, anti-scraping policies employed by websites, inability to find the answer despite multiple attempts etc.). The Oblivious failures are scenarios where Agent-E gives a response that was wrong. These are typically scenarios where agent overlooks certain task requirements and provides an answer that only partially meets the requirements. It could also stem from DOM observation issues (e.g., not being aware that the date got reset due to incorrect format in Google Flights) or website capability understanding issues (e.g., not using advanced search capability when needed, or assuming search functionalities are perfect and every search result will completely satisfy the search requirements). Similar error modes were also observed by \cite{he2024webvoyager} who classify them as agent \textit{hallucinations}.

\subsubsection*{Task Completion Times}
Agent-E on an average took 150 seconds to successfully complete a task and 220 seconds for completion when the task was a failure. The longer duration for failed tasks is interesting and also expected, since given a difficult task, Agent-E may try multiple approaches to complete the task before giving up on it. The difference in task completion times across websites (e.g., 68 seconds to successfully complete a task in WolframAlpha vs. 286 seconds in Amazon) also reflects the differences in task and website complexity. Amazon.com is a feature-rich and content-heavy website (for example, the Amazon and Booking.com homepage contains 3189 and 2,952 DOM elements respectively, in contrast to 1520 in Google Flights and 590 for WolframAlpha homepages). When the website is content-heavy and has numerous DOM elements, the DOM de-noising step takes more time, and the LLM token count for browser navigation agent may increase as well.

\subsubsection*{LLM Calls}
On an average Agent-E took 25 LLM calls to execute a task (6.4 calls by the planner and almost 3 times as much by the browser navigation agent). The average number of LLM calls per website, as expected, is consistent with task completion times. Unfortunately, no prior studies have reported an end-to-end number of LLM calls required for task completion. While \cite{lutz2024wilbur} provides the number of LLM calls required by one (out of three) of the LLM-based components in their system, this information is insufficient for a conclusive comparison.

\subsection{Result: Qualitative}
In this section, we will present qualitative results of concrete examples showing how different design choices made in Agent-E help perform complex web tasks.
\subsubsection*{Hierarchical Planner helps error detection and recovery}
\begin{figure} [h!]
    \centering
    \includegraphics[clip, trim=0 0 0 0, width=\linewidth]{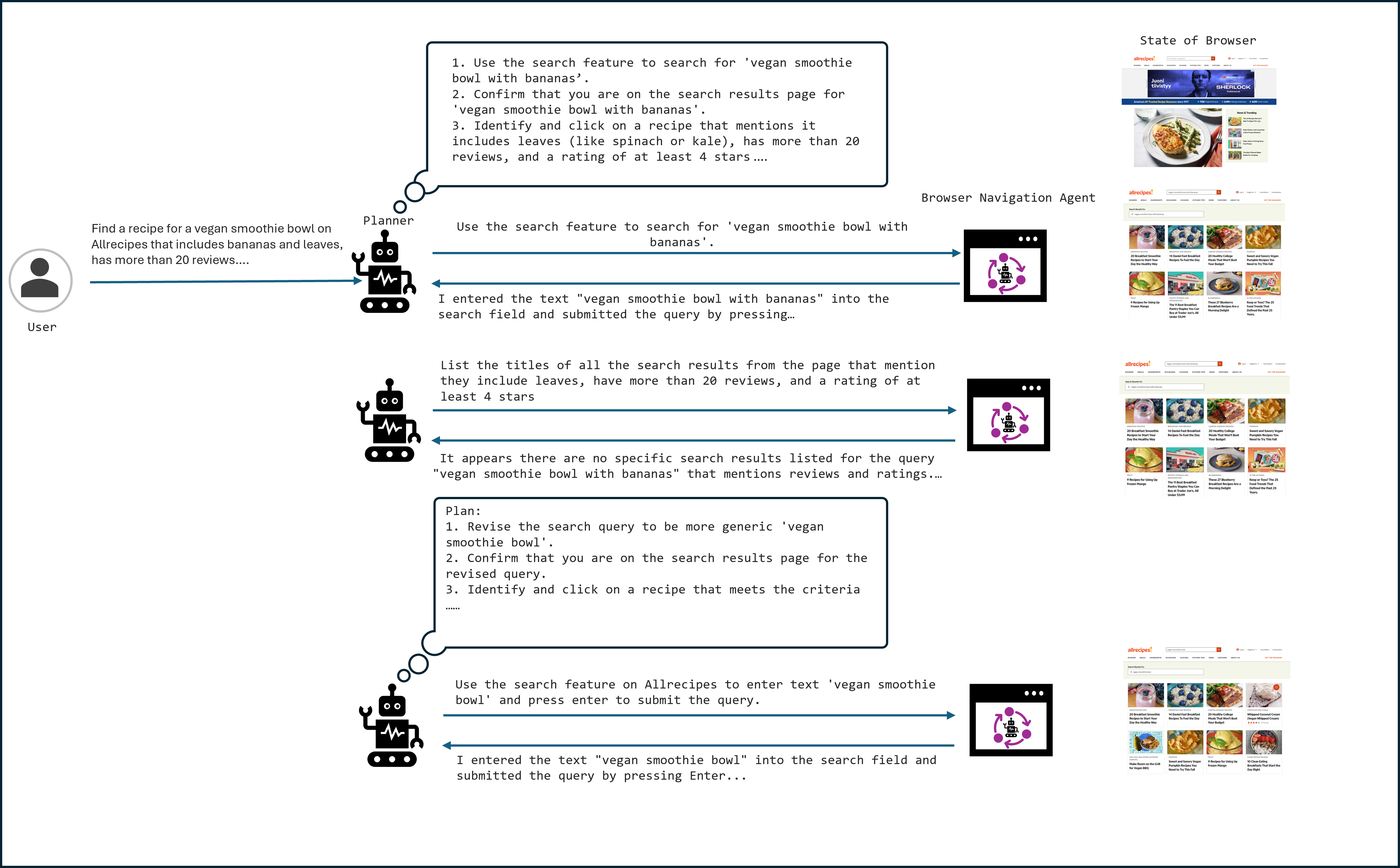}
    \caption{An example instance of Agent-E detecting and recovering from errors. The conversation is truncated with `...' to enhance readability in the image.}
    \label{fig:agente-error}
\end{figure}

The hierarchical architecture allows easily detecting and  recovering from errors. The planner agent is prompted to perform verification (by asking questions or asking confirmation) as part of the plan whenever necessary. Figure \ref{fig:agente-error} shows an example instance where the planner agent asks the browser navigation agent for more information (i.e., \textit{list the search results}), and from the response (i.e., \textit{there are no specific search results}) identifies that it may have made an error by making the search query too focused. In the example, the planner creates a new plan of action for performing the task. 
Another common pattern in the evaluation was the planner's ability to detect errors and easily backtrack to a previous page to continue execution. Given that the planner has the URL of the page at each step available to it, it allows the planner to effortlessly backtrack to a previous page by adding it as a step in the plan (e.g., \textit{navigate to the search result page using the \textless url\textgreater}). 

\subsubsection*{Support for multiple DOM observation methods}
\begin{figure} [h!]
    \centering
    \includegraphics[clip, trim=0 0 0 0, width=\linewidth]{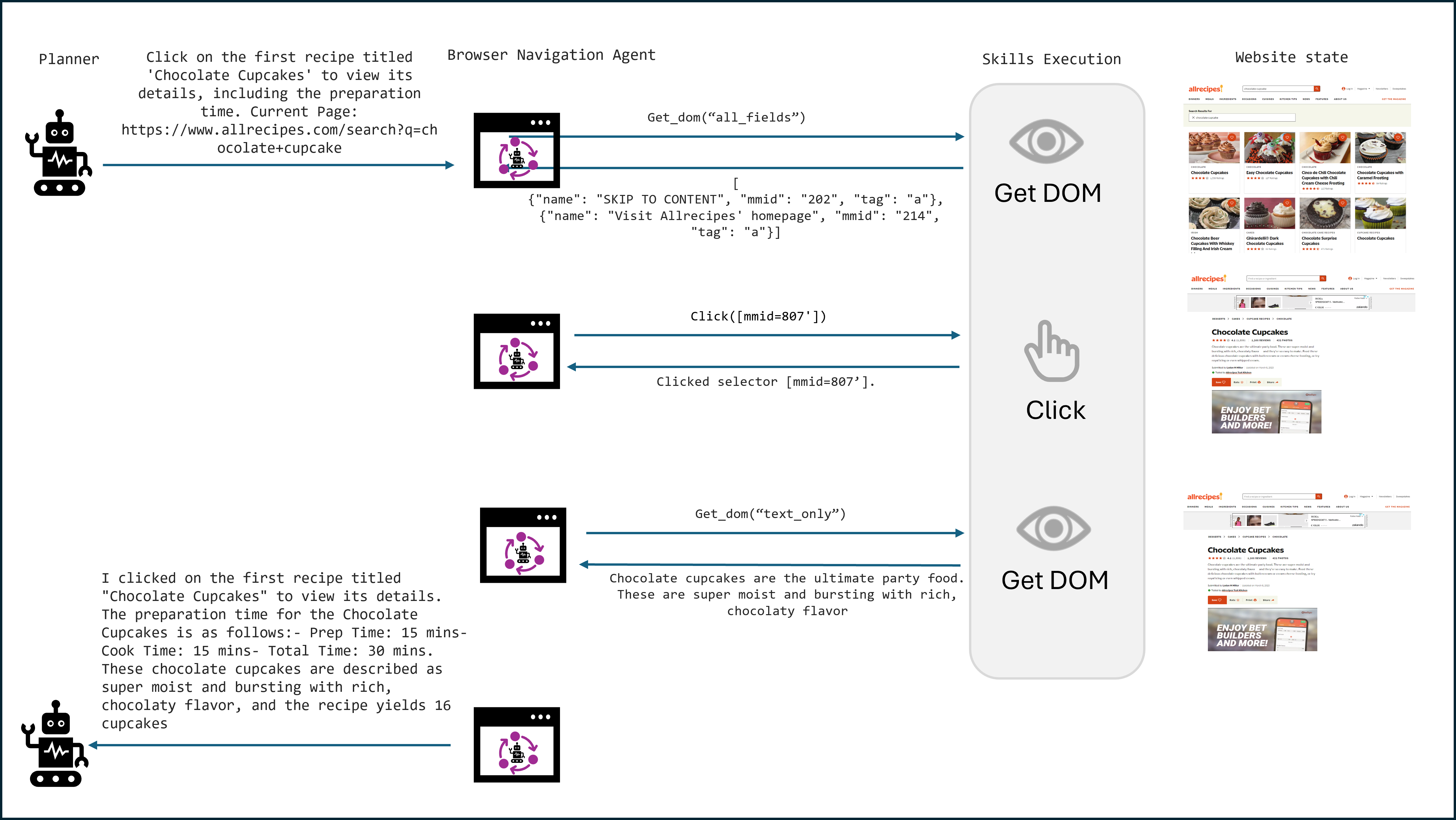}
    \caption{Providing multiple options for DOM observation allows to flexibly select one fit for task. The conversation is truncated with `...' to enhance readability in the image.}
    \label{fig:agente-dom}
\end{figure}

HTML DOM can be extremely large (e.g., YouTube Homepage with all the DOM elements and their attributes is about 800,000 tokens). Thus, it is important to denoise and encode the DOM such that only task relevant information is presented to the LLM. However, information relevant for a given task is very dependent on the task at the hand. Some tasks may only need a complete textual representation (e.g., \textit{summarise the current page}), some tasks may only need input fields and buttons (e.g., \textit{search on google}). On the other hand, more exploratory tasks, may need a complete representation of the page (e.g., \textit{what elements are on this page}). 
Most previous web agent work have used a single DOM representation, e.g. \cite{zhou2023webarena} used accessibility tree, multi-modal web agent in \cite{he2024webvoyager} used screenshots and \cite{lutz2024wilbur} used direct encoding and denoising of the HTML DOM. However, in our view, there is no single DOM observation method that suits all the tasks. Thus, Agent-E supports three different DOM representation methods \textit{text\_only, input\_fields, all\_fields}.  This allows Agent-E to flexibly select the DOM representation that it feels is best suited for the task. Also, it allows for backup, that if one does not work as expected, Agent-E can fall back to a different representation.
There were numerous examples in our benchmark where this multiple DOM representation was useful. Figure \ref{fig:agente-dom} illustrates an example where Agent-E adaptively uses \textit{all\_fields} DOM representation for interaction and \textit{text\_only} for summarization. 
 
\subsubsection*{Change observation helps grounding}
Change observation is a concept we introduced in which each action execution also observes for changes in the state and returns a linguistic feedback to the LLM. Some typical scenarios where this is useful is where the browser navigation agent tries to click on a navigation item (e.g., \textit{click on the soccer link on ESPN.com}), however instead of navigating to the relevant section, it may instead open a popup menu that requires further selection. Another common example is when browser navigation agent attempts to set the source airport in a flight booking websites and a list of possible airports open up as a drop down. In both these cases, the interaction is not yet complete, but the browser navigation agent may assume it is (i.e., navigation to soccer section would require clicking on a new link from the popup menu, setting source airport would require selecting one of the option from the drop down). In both these cases, the \textit{click} skills will return a feedback to the LLM that \textit{as a consequence of the click, a menu has appeared where you may need to make further selection}. See Figure \ref{fig:agente-nested} for an example. 
Conceptually, the purpose of the change observation is to provide linguistic feedback to the LLM whether the action has been completed without any errors and did it lead to any tangible change in the environment, in order to inform subsequent actions. We also envision efficiency improvements if the change observation can return a list of elements, so that LLM can make subsequent selection without again using Get DOM skill to observe the state of the DOM. 
Change observation is adjacent to the concept of Reflexion \cite{shinn2024reflexion}. However, there are nuanced differences between the two. Reflexion paradigm provides feedback of a prior failure. However, change observation is not directly associated or limited to a prior failure. The observation returned can be any outcome of the action.  
Reflexion uses an LLM to analyse the scalar `success/failure' signal from an action and current trajectory to produce the verbal feedback to the agent. Reflexion did not provide any tangible improvement when directly applied to web automation in the Webshop\cite{yao2022webshop} benchmark. On the other hand, change observation is implemented using classical code-based approaches to observe the consequence of an action and  linguistic feedback of actions is implemented using heuristic-based approach to help the system to be better aware of the current state of the environment, and nudge the system towards the correct next action. 

\section{Discussion}
In this section, we will synthesize our learnings from the development and evaluation of Agent-E into a series of agent design principles. We believe these principles can be generalized beyond the domain of web automation. 

\subsection{Agent Design Principles}
\begin{enumerate}
    \item  \textbf{Well crafted set of primitive skills can enable powerful use-cases}:
    A well crafted ensemble of foundational skills can serve as a building block to support more complex functionalities. LLMs can effectively combine these skills to unlock a broad range of usecases. The analysis of the intended use case is crucial to arrive at  the requisite primitive skills. In the case of Agent-E these primitive skills were \textit{click, enter text, get DOM, Open URL} and \textit{Press Keys}. These were only a subset of what a user could  potentially perform on a page (e.g. we did not support \textit{drag, double click, right click, tab management}, etc. We considered the primitive skills we enabled in Agent-E to be enough for vast majority of general web automation tasks. Nonetheless, if we were to specialise Agent-E to work on certain websites where right-click to select a functionality is a prominent interaction pattern, we would need to introduce that as a primitive skill.
    
    \item  \textbf{Consider hierarchical architecture for complex tasks}: 
    Hierarchical architecture can be useful in agents with multiple LLM-based components. It allows execution of more complex tasks through a clear separation of roles and responsibilities. Hierarchical architecture excels in scenarios where tasks can be decomposed into sub-tasks that need to be handled at different levels of granularity. Additionally, it aids in the identification of tasks that can be executed in parallel, potentially leading to performance enhancements. It also supports the development and improvement of various components in isolation.
    \\It is important to keep in mind that hierarchical architecture may not be suitable for all tasks. In case of Agent-E, if all we had to support were, e.g., navigate to specific URLs or perform simple web search, a hierarchical architecture may be over-complicated. In such cases, a much simpler architecture may suffice.  
    
    \item  \textbf{Perform payload denoising when relevant}:
    Mitigating noise in the payload is critical in creating reliable, cost and time efficient agents. Noise could be in the form of unnecessary or irrelevant information, which could lead to incorrect or sub-optimal performance. It is also important to keep in mind that what is considered noise may be dependent on the task.  
    \\Simple techniques such as filtering irrelevant data, transforming the data to an easily consumable format, and focusing on key information can contribute to more accurate decision-making by the agent. This is especially important in environments with a high degree of uncertainty or very large input payload. In the case of Agent-E, we performed denoising on the HTML DOM. Further, we provided multiple DOM observation capabilities that the agent could adaptively use given the task requirements. 
    
    \item \textbf{Provide linguistic feedback of actions:}
    Our exploration into the space of agents capable of performing actions that has tangible consequences suggests that linguistic feedback of actions could be useful to help executor agents to be better aware of the environment and any consequence of the actions (e.g., \textit{a dialog box appeared as a consequence of the click action}). Change observation helps refine the agent's subsequent actions by providing a clear narrative of cause and effect, and also improved awareness of the environment. This could apply in a variety of usecases such as desktop automation or automations in the physical space (e.g robot control). 
    
    \item  \textbf{Support human in the loop architecture when necessary}:
    Supporting human in the loop architecture is critical for agentic systems. Given that agentic systems are so new, users may not trust the system completely. Further, there will arise scenarios where the agent may need to have varying degree of user involvement for a multitude of reasons: e.g., to get more clarification on the task, when there is ambiguity how to proceed, to approve critical actions, perform some mission critical actions that the agent cannot or should not perform, audit the final output and re-invoke if needed, or take over responsibility when agent identifies that it cannot perform the task. Human in the loop architecture could also be leveraged for self-improvement where the agent identifies opportune times to gather human demonstration (e.g. \textit{Sorry, I could not do this task. Can you show me how you would do it?}. Thus, incorporating mechanisms for human oversight can play a pivotal role in building that trust, and seamless handover. 
    \\Human in the loop could mean different levels of human involvement. From simple notifications of task progress, human involvement at key decision points, all the way to a handover of an incomplete task to a human.

    \item \textbf{Analyse, reflect and aggregate past experiences routinely for self-improvement:}
    For Agentic systems to be adopted widely, they need to (gradually) achieve close to human-level performance. In the context of web automation, Agent-E could perform 73.1\% of the tasks with an average task completion time of 150-220 second and 25 LLM calls. While very promising, this is not practical for production use-cases in terms of both effectiveness and efficiency.
    \\A simple agentic solution to improving efficiency is to cache LLM calls. These mechanisms are supported out-of-the-box by agent development frameworks such as Autogen \cite{wu2023autogen}. However, a naive caching may not work well in dynamic contexts (e.g., a web site will continuously change in terms of content, advertisements etc.). A better approach would be to establish offline workflows that routinely analyse, reflects on and aggregates past tasks and human demonstrations to convert them to more classical automation workflows. These classical approach could be re-triggered upon a new task if it matches a workflow that it has encountered in the past, and use the exploratory agentic approach only as a fallback. This would mean the task could be completed faster and cheaper.  

    \item \textbf{Introduce internal and external guardrails into the Agentic System:} To ensure safe and effective operation of an agentic system, it's important to introduce various guardrails into its operation. These guardrails can be in the form of rules, guidelines or constitution that the system must adhere to. 
    Some of the guardrails may be external to system, e.g., a task appropriateness guardrail may filter out inappropriate tasks even before agentic system is invoked. 
    In addition, there could be guardrails intrinsic to the system such as guardrails against prompt injection attacks, and operational guardrails that limits how the agent performs different tasks. 
    \\In the context of Agent-E, the operational guardrails include domain boundaries within which Agent-E should operate (e.g., \textit{intranet of an organisation}), task specific boundaries (e.g., \textit{Agent-E should only respond to a medical question from authoritative websites}) or user-configured boundaries (e.g., \textit{for shopping, use only Amazon}) 

    \item \textbf{Choose between generic agent vs. task specific agent:}
    Generic agentic systems by definition  can perform a wide range of tasks. However, in many practical implementations, a more focused set of capability may be desirable. 
    For example, Agent-E is a generic web agent that can perform a wide range of tasks on the internet, but not necessarily optimized for any specific task. It would be possible to optimize Agent-E for specific type of tasks (e.g., form filling) or specific websites (e.g., Atlassian Confluence pages) to achieve significantly higher performance. Depending on the use case, an optimized agent may suit better for certain workflows than a generic version.
    
\end{enumerate}

\section{Related Work}

\noindent\textbf{LLM-based Planning and Reasoning} Over the last few years, large language models (LLMs) have demonstrated extraordinary abilities in text generation, code generation, and the generation of natural language multi-step reasoning traces. Spurred by these, there has been much work in the use of LLMs to solve multi-step reasoning and planning problems. The many variants of `chain-of-thought' techniques \cite{wei2023chainofthought, chu2023survey} encourage the LLM to produce a series of tokens with causal decoding that drive towards the solution of problems in math, common-sense reasoning and other similar tasks \cite{chowdhery2022palm, fu2023specializing, li2023textbooks, mitra2024orcamath}. With tool-usage for sensing and acting, LLMs have also been used to drive planning in software environments and embodied agents e.g., \cite{baker2022video, wang2023voyager, wang2023jarvis1, irpan2022saycan, bousmalis2023robocat, wu2023unleashing, bhateja2023robotic}. Finally, there has been related work investigating the limits of LLMs when it comes to planning and validation. For examples of negative results, see \cite{valmeekam2023planning, momennejad2023evaluating, valmeekam2023large, huang2023large, kambhampati2024llmscantplan} among others.
\\ \  \\

\noindent\textbf{Specialized Agents for Repetitive Tasks} Beyond the examples above, and as described in Section~\ref{sec:intro}, there has been much recent interest in building specialized agents for the web \cite{zheng2024gpt, he2024webvoyager, lutz2024wilbur} and on device \cite{bai2024digirl, wen2024autodroid}. Also related is recent work on building agentic workflows to replace robotic process automation \cite{wornow2024eclair}. Further afield, the work on building agents and training language models for API usage is also related, given that many software tasks and workflows involve the use of APIs; examples include \cite{hosseini2021compositional, patil2023gorilla, qin2023toolllm} and many more.
\\ \  \\

\noindent\textbf{Hierarchical Planning} The notion of hierarchical AI planning has been around for five decades or more. Instead of planning directly in the space of low-level primitive actions, planning in a space of `high-level actions' constrains the size of the plan length (and hence the size of the plan-space), which can result in more effective and efficient search. Examples from prior work include \cite{tate1977hierarchy, nau1999shop, marthi2007angelic} and many more; see  \cite{russell2016artificial} for more details. Also related is the use of temporal abstractions in planning and in reinforcement learning, for example the use of options in \cite{sutton1999option, bacon2016optioncritic}. In recent years, multiple papers have proposed the use of hierarchical planning for solving tasks in complex environments or with embodied agents; examples include \cite{wang2023deps, irpan2022saycan} and several others.

\section{Conclusion}
In this paper, we introduced Agent-E, a novel web agent designed to perform complex web-based tasks that makes use of numerous architectural improvements over prior state-of-the-art web agents such as hierarchical architecture, flexible DOM distillation and denoising method and concept of \textit{change observation} to guide the agent towards more accurate performance. 

Agent-E was evaluated on the WebVoyager benchmark, achieving a state-of-the-art success rate of 73.2\%, a significant improvement over previous text-only and multi-modal web agents. Beyond task success rates, we also reported on additional metrics such as error awareness, task completion times, and the number of LLM calls, providing a more comprehensive evaluation of Agent-E's performance.

We presented our learnings in the form of eight general design principles for developing agentic systems that can be applied beyond the realm of web automation. 
\bibliographystyle{apalike}
\bibliography{sample}

\begin{thebibliography}{}

\bibitem[Bacon et~al., 2017]{bacon2016optioncritic}
Bacon, P.-L., Harb, J., and Precup, D. (2017).
\newblock The option-critic architecture.
\newblock {\em AAAI Conference on Artificial Intelligence}.

\bibitem[Bai et~al., 2024]{bai2024digirl}
Bai, H., Zhou, Y., Cemri, M., Pan, J., Suhr, A., Levine, S., and Kumar, A. (2024).
\newblock Digirl: Training in-the-wild device-control agents with autonomous reinforcement learning.
\newblock {\em arXiv preprint arXiv:2406.11896}.

\bibitem[Baker et~al., 2022]{baker2022video}
Baker, B., Akkaya, I., Zhokhov, P., Huizinga, J., Tang, J., Ecoffet, A., Houghton, B., Sampedro, R., and Clune, J. (2022).
\newblock Video pretraining (vpt): Learning to act by watching unlabeled online videos.

\bibitem[Bhateja et~al., 2023]{bhateja2023robotic}
Bhateja, C., Guo, D., Ghosh, D., Singh, A., Tomar, M., Vuong, Q., Chebotar, Y., Levine, S., and Kumar, A. (2023).
\newblock Robotic offline rl from internet videos via value-function pre-training.

\bibitem[Bousmalis et~al., 2023]{bousmalis2023robocat}
Bousmalis, K., Vezzani, G., Rao, D., Devin, C., Lee, A.~X., Bauza, M., Davchev, T., Zhou, Y., Gupta, A., Raju, A., Laurens, A., Fantacci, C., Dalibard, V., Zambelli, M., Martins, M., Pevceviciute, R., Blokzijl, M., Denil, M., Batchelor, N., Lampe, T., Parisotto, E., Żołna, K., Reed, S., Colmenarejo, S.~G., Scholz, J., Abdolmaleki, A., Groth, O., Regli, J.-B., Sushkov, O., Rothörl, T., Chen, J.~E., Aytar, Y., Barker, D., Ortiz, J., Riedmiller, M., Springenberg, J.~T., Hadsell, R., Nori, F., and Heess, N. (2023).
\newblock Robocat: A self-improving foundation agent for robotic manipulation.

\bibitem[Chowdhery et~al., 2022]{chowdhery2022palm}
Chowdhery, A., Narang, S., Devlin, J., Bosma, M., Mishra, G., Roberts, A., Barham, P., Chung, H.~W., Sutton, C., Gehrmann, S., Schuh, P., Shi, K., Tsvyashchenko, S., Maynez, J., Rao, A., Barnes, P., Tay, Y., Shazeer, N., Prabhakaran, V., Reif, E., Du, N., Hutchinson, B., Pope, R., Bradbury, J., Austin, J., Isard, M., Gur-Ari, G., Yin, P., Duke, T., Levskaya, A., Ghemawat, S., Dev, S., Michalewski, H., Garcia, X., Misra, V., Robinson, K., Fedus, L., Zhou, D., Ippolito, D., Luan, D., Lim, H., Zoph, B., Spiridonov, A., Sepassi, R., Dohan, D., Agrawal, S., Omernick, M., Dai, A.~M., Pillai, T.~S., Pellat, M., Lewkowycz, A., Moreira, E., Child, R., Polozov, O., Lee, K., Zhou, Z., Wang, X., Saeta, B., Diaz, M., Firat, O., Catasta, M., Wei, J., Meier-Hellstern, K., Eck, D., Dean, J., Petrov, S., and Fiedel, N. (2022).
\newblock Palm: Scaling language modeling with pathways.

\bibitem[Chu et~al., 2023]{chu2023survey}
Chu, Z., Chen, J., Chen, Q., Yu, W., He, T., Wang, H., Peng, W., Liu, M., Qin, B., and Liu, T. (2023).
\newblock A survey of chain of thought reasoning: Advances, frontiers and future.

\bibitem[Deng et~al., 2024]{deng2024mind2web}
Deng, X., Gu, Y., Zheng, B., Chen, S., Stevens, S., Wang, B., Sun, H., and Su, Y. (2024).
\newblock Mind2web: Towards a generalist agent for the web.
\newblock {\em Advances in Neural Information Processing Systems}, 36.

\bibitem[Fu et~al., 2023]{fu2023specializing}
Fu, Y., Peng, H., Ou, L., Sabharwal, A., and Khot, T. (2023).
\newblock Specializing smaller language models towards multi-step reasoning.

\bibitem[He et~al., 2024]{he2024webvoyager}
He, H., Yao, W., Ma, K., Yu, W., Dai, Y., Zhang, H., Lan, Z., and Yu, D. (2024).
\newblock Webvoyager: Building an end-to-end web agent with large multimodal models.
\newblock {\em arXiv preprint arXiv:2401.13919}.

\bibitem[Hosseini et~al., 2021]{hosseini2021compositional}
Hosseini, S., Awadallah, A.~H., and Su, Y. (2021).
\newblock Compositional generalization for natural language interfaces to web apis.
\newblock {\em arXiv preprint arXiv:2112.05209}.

\bibitem[Huang et~al., 2023]{huang2023large}
Huang, J., Chen, X., Mishra, S., Zheng, H.~S., Yu, A.~W., Song, X., and Zhou, D. (2023).
\newblock Large language models cannot self-correct reasoning yet.

\bibitem[Irpan et~al., 2022]{irpan2022saycan}
Irpan, A., Herzog, A., Toshev, A.~T., Zeng, A., Brohan, A., Ichter, B.~A., David, B., Parada, C., Finn, C., Tan, C., Reyes, D., Kalashnikov, D., Jang, E.~V., Xia, F., Rettinghouse, J.~L., Hsu, J.~C., Quiambao, J.~L., Ibarz, J., Rao, K., Hausman, K., Gopalakrishnan, K., Lee, K.-H., Jeffrey, K.~A., Luu, L., Yan, M., Ahn, M.~S., Sievers, N., Joshi, N.~J., Brown, N., Cortes, O. E.~E., Xu, P., Sampedro, P.~P., Sermanet, P., Ruano, R.~J., Julian, R.~C., Jesmonth, S.~A., Levine, S., Xu, S., Xiao, T., Vanhoucke, V.~O., Lu, Y., Chebotar, Y., and Kuang, Y. (2022).
\newblock Do as i can, not as i say: Grounding language in robotic affordances.
\newblock {\em arXiv preprint arXiv:2204.01691}.

\bibitem[Kambhampati et~al., 2024]{kambhampati2024llmscantplan}
Kambhampati, S., Valmeekam, K., Guan, L., Verma, M., Stechly, K., Bhambri, S., Saldyt, L., and Murthy, A. (2024).
\newblock Llms can't plan, but can help planning in llm-modulo frameworks.
\newblock {\em arXiv preprint arXiv:2402.01817}.

\bibitem[Kapoor et~al., 2024]{kapoor2024ai}
Kapoor, S., Stroebl, B., Siegel, Z.~S., Nadgir, N., and Narayanan, A. (2024).
\newblock Ai agents that matter.
\newblock {\em arXiv preprint arXiv:2407.01502}.

\bibitem[Li et~al., 2023]{li2023textbooks}
Li, Y., Bubeck, S., Eldan, R., Giorno, A.~D., Gunasekar, S., and Lee, Y.~T. (2023).
\newblock Textbooks are all you need ii: phi-1.5 technical report.

\bibitem[Lutz et~al., 2024]{lutz2024wilbur}
Lutz, M., Bohra, A., Saroyan, M., Harutyunyan, A., and Campagna, G. (2024).
\newblock Wilbur: Adaptive in-context learning for robust and accurate web agents.
\newblock {\em arXiv preprint arXiv:2404.05902}.

\bibitem[Marthi et~al., 2007]{marthi2007angelic}
Marthi, B., Russell, S., and Wolfe, J. (2007).
\newblock Angelic semantics for high-level actions.
\newblock {\em International Conference on Automated Planning and Scheduling}.

\bibitem[Mitra et~al., 2024]{mitra2024orcamath}
Mitra, A., Khanpour, H., Rosset, C., and Awadallah, A. (2024).
\newblock Orca-math: Unlocking the potential of slms in grade school math.

\bibitem[Momennejad et~al., 2023]{momennejad2023evaluating}
Momennejad, I., Hasanbeig, H., Vieira, F., Sharma, H., Ness, R.~O., Jojic, N., Palangi, H., and Larson, J. (2023).
\newblock Evaluating cognitive maps and planning in large language models with cogeval.

\bibitem[Nakano et~al., 2022]{nakano2022webgpt}
Nakano, R., Hilton, J., Balaji, S., Wu, J., Ouyang, L., Kim, C., Hesse, C., Jain, S., Kosaraju, V., Saunders, W., Jiang, X., Cobbe, K., Eloundou, T., Krueger, G., Button, K., Knight, M., Chess, B., and Schulman, J. (2022).
\newblock Webgpt: Browser-assisted question-answering with human feedback.
\newblock {\em arXiv preprint arXiv:2112.09332}.

\bibitem[Nau et~al., 1991]{nau1999shop}
Nau, D., Cao, Y., Lotem, A., and Muñoz-Avila, H. (1991).
\newblock Shop: Simple hierarchical ordered planner.
\newblock {\em International Joint Conference on Artificial Intelligence}.

\bibitem[Patil et~al., 2023]{patil2023gorilla}
Patil, S.~G., Zhang, T., Wang, X., and Gonzalez, J.~E. (2023).
\newblock Gorilla: Large language model connected with massive apis.
\newblock {\em arXiv preprint arXiv:2305.15334}.

\bibitem[Qin et~al., 2024]{qin2023toolllm}
Qin, Y., Liang, S., Ye, Y., Zhu, K., Yan, L., Lu, Y., Lin, Y., Cong, X., Tang, X., Qian, B., Zhao, S., Hong, L., Tian, R., Xie, R., Zhou, J., Gerstein, M., Li, D., Liu, Z., and Sun, M. (2024).
\newblock Toolllm: Facilitating large language models to master 16000+ real-world apis.
\newblock {\em International Conference on Learning Representations}.

\bibitem[Russell and Norvig, 2009]{russell2016artificial}
Russell, S.~J. and Norvig, P. (2009).
\newblock {\em Artificial Intelligence: a modern approach}.
\newblock Pearson, 3 edition.

\bibitem[Shinn et~al., 2024]{shinn2024reflexion}
Shinn, N., Cassano, F., Gopinath, A., Narasimhan, K., and Yao, S. (2024).
\newblock Reflexion: Language agents with verbal reinforcement learning.
\newblock {\em Advances in Neural Information Processing Systems}, 36.

\bibitem[Sutton et~al., 1999]{sutton1999option}
Sutton, R., Precup, D., and Singh, S. (1999).
\newblock Between mdps and semi-mdps: A framework for temporal abstraction in reinforcement learning.
\newblock {\em Artificial Intelligence Journal}.

\bibitem[Tate, 1977]{tate1977hierarchy}
Tate, A. (1977).
\newblock Generating project networks.
\newblock {\em International Joint Conference on Artificial Intelligence}.

\bibitem[Valmeekam et~al., 2023a]{valmeekam2023large}
Valmeekam, K., Marquez, M., and Kambhampati, S. (2023a).
\newblock Can large language models really improve by self-critiquing their own plans?

\bibitem[Valmeekam et~al., 2023b]{valmeekam2023planning}
Valmeekam, K., Sreedharan, S., Marquez, M., Olmo, A., and Kambhampati, S. (2023b).
\newblock On the planning abilities of large language models (a critical investigation with a proposed benchmark).

\bibitem[Wang et~al., 2023a]{wang2023voyager}
Wang, G., Xie, Y., Jiang, Y., Mandlekar, A., Xiao, C., Zhu, Y., Fan, L., and Anandkumar, A. (2023a).
\newblock Voyager: An open-ended embodied agent with large language models.

\bibitem[Wang et~al., 2022]{wang2023deps}
Wang, Z., Cai, S., Chen, G., Liu, A., Ma, X., and Liang, Y. (2022).
\newblock Describe, explain, plan and select: Interactive planning with large language models enables open-world multi-task agents.
\newblock {\em Advances in Neural Information Processing Systems}, 37.

\bibitem[Wang et~al., 2023b]{wang2023jarvis1}
Wang, Z., Cai, S., Liu, A., Jin, Y., Hou, J., Zhang, B., Lin, H., He, Z., Zheng, Z., Yang, Y., Ma, X., and Liang, Y. (2023b).
\newblock Jarvis-1: Open-world multi-task agents with memory-augmented multimodal language models.

\bibitem[Wei et~al., 2023]{wei2023chainofthought}
Wei, J., Wang, X., Schuurmans, D., Bosma, M., Ichter, B., Xia, F., Chi, E., Le, Q., and Zhou, D. (2023).
\newblock Chain-of-thought prompting elicits reasoning in large language models.

\bibitem[Wen et~al., 2024]{wen2024autodroid}
Wen, H., Li, Y., Liu, G., Zhao, S., Yu, T., Li, T. J.-J., Jiang, S., Liu, Y., Zhang, Y., and Liu, Y. (2024).
\newblock Autodroid: Llm-powered task automation in android.
\newblock In {\em Proceedings of the 30th Annual International Conference on Mobile Computing and Networking}, pages 543--557.

\bibitem[Wornow et~al., 2024]{wornow2024eclair}
Wornow, M., Narayan, A., Opsahl-Ong, K., McIntyre, Q., Shah, N.~H., and Re, C. (2024).
\newblock Automating the enterprise with foundation models.
\newblock {\em arXiv preprint arXiv:2405.03710}.

\bibitem[Wu et~al., 2023a]{wu2023unleashing}
Wu, H., Jing, Y., Cheang, C., Chen, G., Xu, J., Li, X., Liu, M., Li, H., and Kong, T. (2023a).
\newblock Unleashing large-scale video generative pre-training for visual robot manipulation.

\bibitem[Wu et~al., 2023b]{wu2023autogen}
Wu, Q., Bansal, G., Zhang, J., Wu, Y., Zhang, S., Zhu, E., Li, B., Jiang, L., Zhang, X., and Wang, C. (2023b).
\newblock Autogen: Enabling next-gen llm applications via multi-agent conversation framework.
\newblock {\em arXiv preprint arXiv:2308.08155}.

\bibitem[Yao et~al., 2022]{yao2022webshop}
Yao, S., Chen, H., Yang, J., and Narasimhan, K. (2022).
\newblock Webshop: Towards scalable real-world web interaction with grounded language agents.
\newblock {\em Advances in Neural Information Processing Systems}, 35:20744--20757.

\bibitem[Zheng et~al., 2024]{zheng2024gpt}
Zheng, B., Gou, B., Kil, J., Sun, H., and Su, Y. (2024).
\newblock Gpt-4v (ision) is a generalist web agent, if grounded.
\newblock {\em arXiv preprint arXiv:2401.01614}.

\bibitem[Zhou et~al., 2023]{zhou2023webarena}
Zhou, S., Xu, F.~F., Zhu, H., Zhou, X., Lo, R., Sridhar, A., Cheng, X., Bisk, Y., Fried, D., Alon, U., et~al. (2023).
\newblock Webarena: A realistic web environment for building autonomous agents.
\newblock {\em arXiv preprint arXiv:2307.13854}.

\end{thebibliography}

\appendix 

\end{document}